\documentclass[ijds,sglanonrev]{informs4}
\usepackage{eqndefns-left} 
\RequirePackage{tgtermes}
\RequirePackage{newtxtext}
\RequirePackage{newtxmath}
\RequirePackage{bm}
\RequirePackage{endnotes}

\OneAndAHalfSpacedXII 

\usepackage{algorithm}
\usepackage{algpseudocode}
\usepackage{tikz}

\usepackage{natbib}
 \bibpunct[, ]{(}{)}{,}{a}{}{,}%
 \def\bibfont{\small}%
\EquationsNumberedThrough    

\TheoremsNumberedThrough     
\ECRepeatTheorems  %

\MANUSCRIPTNO{IJDS-0001-2024.00}

\usepackage{hyperref}

\begin{document}


\RUNAUTHOR{Chokhachian, Katzfuss, and Ding}

\RUNTITLE{Fast Gaussian Process Approximations for Autocorrelated Data}

\TITLE{Fast Gaussian Process Approximations for Autocorrelated Data}

\ARTICLEAUTHORS{%
\AUTHOR{Ahmadreza Chokhachian\textsuperscript{a}, Matthias Katzfuss\textsuperscript{b}, and Yu Ding\textsuperscript{c*}}
\AFF{\textsuperscript{a,c}H. Milton Stewart School of Industrial and Systems Engineering, Georgia Institute of Technology, Atlanta, GA 30339}
\AFF{\textsuperscript{b}Department of Statistics, University of Wisconsin--Madison, Madison, WI 53706}

\AFF{\textsuperscript{*}Corresponding author}
{\small\vspace{6pt}
\noindent\textbf{Contact:} chokhachian@gatech.edu (A-C); mkatzfuss@wisc.edu (M-K); yu.ding@isye.gatech.edu (Y-D)}
}

\ABSTRACT{%
This paper is concerned with the problem of how to speed up computation for Gaussian process models trained on autocorrelated data. The Gaussian process model is a powerful tool commonly used in nonlinear regression applications. Standard regression modeling assumes random samples and an independently, identically distributed noise. Various fast approximations that speed up Gaussian process regression work under this standard setting.  But for autocorrelated data, failing to account for autocorrelation leads to a phenomenon known as temporal overfitting that deteriorates model performance on new test instances. To handle autocorrelated data, existing fast Gaussian process approximations have to be modified; one such approach is to segment the originally correlated data points into blocks in which the blocked data are de-correlated. This work explains how to make some of the existing Gaussian process approximations work with blocked data. Numerical experiments across diverse application datasets demonstrate that the proposed approaches can remarkably accelerate computation for Gaussian process regression on autocorrelated data without compromising model prediction performance. 
}%

\FUNDING{This work was supported in part by the NSF (Grants CNS--2328395, DMS--1953005), the Wisconsin Alumni Research Foundation, and the Ocean Energy Safety Institute Consortium (OESIC).}




\maketitle

\section{Introduction\label{sec:intro}}
A standard nonlinear regression model assumes a relationship between the response \( y \) and input vector \( \boldsymbol{x} \), typically expressed as:
\begin{equation}
    y_i = f(\boldsymbol{x}_i) + \epsilon_i,
\end{equation}
where \( f(\cdot) \) is the unknown function to be estimated.  Both $\boldsymbol{x}$ and $y$ are drawn randomly from their joint distribution and \( \epsilon \) represents a Gaussian noise, assumed to be identically and independently distributed (i.i.d). Gaussian process (GP) regression is a popular modeling tool used for nonlinear regression \citep{Rasmussen06}, in which \( f(\cdot) \) is modeled as a realization of a GP by specifying its mean and covariance functions. The model parameters are trained using \( n \) pairs of training data points, which we denote by \( D_i = \{ \boldsymbol{x}_i, y_i \} \), $i=1,\ldots,n$.

A known issue with GP regression is the \( O(n^3) \) computational complexity needed for inverting the covariance matrices of size $n \times n$, which slows down the model fitting considerably when working on large-scale datasets (say, $n>10{,}000$). Various approximation methods have been developed to address this computational bottleneck \citep[e.g.,][]{Heaton2017}. We consider here three popular and broadly used methods, namely local approximate Gaussian process \citep[laGP]{Gramacy2015}, twin Gaussian process \citep[twinGP]{Vakayil2024}, and Vecchia approximations \citep{katzfuss2022}.  These approximations do speed up the GP computation remarkably and shorten model fitting from several hours to minutes or even seconds.  But their target application is on the standard regression models for which random sampling and i.i.d.\ noise are assumed. 

For more than a decade, a series of studies \citep{Xiao2003, Sheridan2013, Roberts2017, Meyer2018, Rabinowicz2020, Prakash2023} pointed out that the i.i.d.\ assumption for $\epsilon$ does not always hold and that $\boldsymbol{x}$ and $y$ are not drawn randomly, especially for data coming out of physical and engineering systems. Autocorrelation in the physical world is caused mainly, but not exclusively, by $\boldsymbol x$. Take wind power production as an example. The input $\boldsymbol x$ includes wind speed, which is the driving force behind wind power production.  Because wind speed is autocorrelated (the mass of air has inertia), the power output $y$ is also autocorrelated.  But $y$ is affected by more than wind speed and not all of the influencing factors are modeled in $\boldsymbol x$. As such, even after $\boldsymbol x$ is de-autocorrelated, $y$ could still have some residual autocorrelation. 

Ignoring the autocorrelation in data distorts model selection and compromises parameter estimation, leading to a phenomenon known as \emph{temporal overfitting} and deteriorating the model performance on future data. The aforementioned series of studies developed several approaches to handle the autocorrelated data and avoid temporal overfitting. Among them, \citet{Prakash2023} proposed a modified GP model, nicknamed \emph{tempGP}, that can effectively model autocorrelated data for nonlinear regression and outperform other methods developed prior to tempGP. Being a GP model, tempGP naturally suffers from the same computational inefficiency described above.  For example, when tempGP is used to fit a dataset of $50{,}000$ data points, it takes more than 3 hours on high-performance computers, whereas a fast GP method may take just a few minutes for fitting a dataset of the same size.  

There is a clear need to speed up tempGP for handling large autocorrelated data, but applying existing GP approximation methods directly to autocorrelated data is not desirable. The reason is as follows. The three existing GP approximation methods mentioned above all rely on the use of the nearest neighbors defined by $\boldsymbol x$, but using the nearest neighbors induces temporal proximity, reinforces autocorrelation, and exacerbates temporal overfitting in prediction. In other words, using the existing GP approximations on autocorrelated data, while doing so can speed up computation, suffers from poor performance.  

Since tempGP avoids temporal overfitting, one wonders if existing GP approximation methods can be used directly with tempGP. The answer is ``yes'' but some modification is needed, so that the computation in tempGP can be sped up without sacrificing model performance on autocorrelated data. This paper explores strategies that make three main existing GP approximations, laGP, twinGP, and scaled Vecchia (SV), to work with the special treatment introduced in tempGP, known as data thinning. As such, the resulting GP approximations for autocorrelated data will be called \emph{thinned laGP}, \emph{thinned twinGP}, and \emph{thinned SV}, respectively. Our extensive numerical experiments show that thinned SV has the overall best model performance on autocorrelated data, whereas thinned twinGP is the fastest approximation.  Thinned laGP does not appear to be competitive as compared with the other two alternatives.

The rest of the paper unfolds as follows.  Section 2 briefly recaps the model and data thinning used in tempGP for handling autocorrelated data. Section 3 explains how the existing GP approximations work with data thinning. Section 4 presents a series of numerical experiments that evaluate which alternatives are most advantageous. Finally, Section 5 concludes the paper.

\section{TempGP and data thinning}
\label{overview}

Let us first take a look at the treatment in tempGP for handling autocorrelated data and avoiding temporal overfitting. \citet{Prakash2023} incorporate a time index \( t_i \), which is the time when the data point $\{ \boldsymbol{x}_i, y_i\}$ is recorded, into the representation of the data point. As such, the data point is then expressed as \( D_i = \{ \boldsymbol{x}_i, y_i, t_i \} \) and data points are arranged following the natural time sequence. If the raw data are not arranged following the time sequence, they can be easily rearranged, given the information of $t_i$. Understandably, for some datasets that have autocorrelation among the data points, they may lose the time information $t_i$; for instance, the time stamps were not recorded in the raw data. For this kind of datasets, one can let $t_i=i$. Such treatment is not ideal, because doing so makes the data points evenly spaced along the time axis but there is no guarantee that is always true in the raw data. Thankfully, for many engineering system, automated data collection mechanisms like inline sensors do churn out data sequences with evenly spaced data points.  That explains why such a treatment tends to work for most applications. 

\begin{figure}[ht]
\centering
\includegraphics[width=\textwidth]{"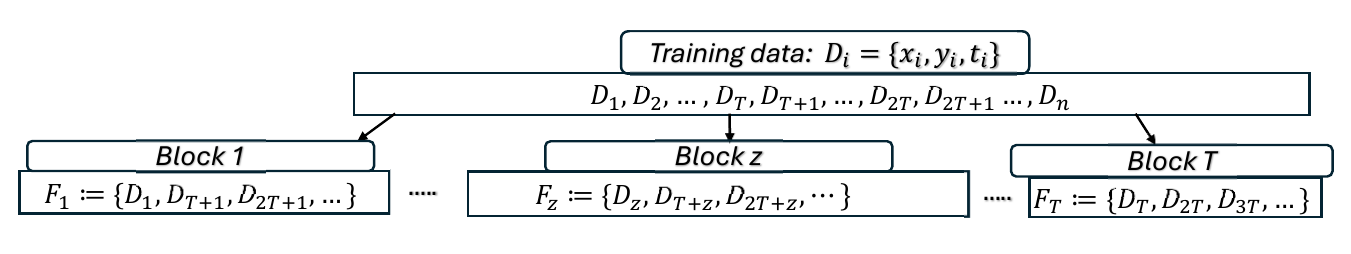"}
\caption{The thinning mechanism}
\label{fig:thinning}
\end{figure}

With the new expression of $D_i$, \citet{Prakash2023} introduce the following model:
\begin{equation}
\label{Model2}
    y_i = f(\boldsymbol{x}_i) + g(t_i) + \epsilon_i,
\end{equation}
which uses \( g(t_i) \) to capture the temporally autocorrelated component, so that  \( \epsilon_i \) can still be i.i.d. In this model, $\boldsymbol{x}_i$ can be autocorrelated but the function relationship $f(\cdot)$ is assumed to be time invariant.  \citet{Prakash2023} model both \( f(\cdot) \) and \( g(\cdot) \) as GPs (thus the name tempGP) and devise a two-step estimation process to estimate \( f(\cdot) \) and \( g(\cdot) \), respectively. Specifically, \citet{Prakash2023} estimate $f(\boldsymbol x_i)$ first by ignoring the $g(t_i)$ term. For doing that, \citet{Prakash2023} thin the original time sequenced data into $T$ blocks, as illustrated in Figure \ref{fig:thinning}, so that the data points in each block $F_z$ have a much greater time interval in between.  The thinning number $T$  is chosen such that the data points in each block become uncorrelated.  The process resembles the thinning process used in the Markov chain Monte Carlo (MCMC) sampling. Because the data points in each block $F_z$ are uncorrelated, the $g(t_i)$ term naturally disappears from Model (\ref{Model2}) when the model is applied over $F_z$.  \citet{Prakash2023} define a pseudo-likelihood function as a joint product over the $T$ blocks of uncorrelated data for estimating $f(\boldsymbol x_i)$.  Once $f(\boldsymbol x_i)$ is estimated, \citet{Prakash2023} use the residuals, i.e., $y_i - \hat{f}(\boldsymbol x_i)$, to estimate $g(t_i)$.

\citet{Prakash2023} set $T$ so that the autocorrelation in $\boldsymbol{x}$ becomes weak enough: 
\begin{equation}
    \label{thinning_x}
    T =  \arg\min_h \max_{\ell=1, \ldots, d} \left(\left| \operatorname{PACF}_{(\boldsymbol{x})_{\ell}}(h) \right| \leq \frac{2}{\sqrt{n}}\right),
\end{equation}
where $\operatorname{PACF}_{(\boldsymbol{x})_{\ell}}(h)$ is the partial autocorrelation function for the $\ell$-th covariate for lag $h$. The threshold is chosen to be $\frac{2}{\sqrt{n}}$ as approximately the 95\% confidence interval under the null hypothesis of no autocorrelation. With such a setup, tempGP outperforms several competitive methods by an appreciable margin for handling autocorrelated data: the Pre-Whitening approach~\citep{Xiao2003}, the time-split cross-validation~\citep{Sheridan2013, Roberts2017, Meyer2018}, and the corrected cross-validation~\citep{Rabinowicz2020}.  

If one desires no autocorrelation in both covariates $\boldsymbol{x}$ and response $y$, it would be better to include $y$ in the PACF assessment, such as
\begin{equation}
\label{thinning_x&y}
T = \arg\min_{h}\;\max_{\,z\in\{x_1,\dots,x_\ell, y\}}
\Bigl(\bigl|\mathrm{PACF}_{(z)}(h)\bigr|\le\tfrac{2}{\sqrt{n}}\Bigr).
\end{equation}
In this paper, we decide to use Equation~(\ref{thinning_x&y}) to determine $T$.  The outcomes are rather close, and sometimes identical, to the $T$ found using Equation~(\ref{thinning_x}).  The reason is simple, for the autocorrelation in $y$ is mainly driven by the autocorrelation in $\boldsymbol{x}$ and therefore once $\boldsymbol{x}$ is uncorrelated, the autocorrelation in $y$ will also be significantly weakened. 

\section{GP approximations for autocorrelated data}
The question is whether the existing GP approximation methods can be used to speed up tempGP. The short answer is yes, but certain modifications have to be made to ensure existing GP approximations work with the data thinning scheme.

Before we explain how to modify GP approximations to work with thinned data, we would first like to articulate how the neighborhood in a dataset is defined, because all of the three GP approximations described in Section~\ref{sec:intro} make use of some neighborhood of data (i.e., a small subset of data) for reducing computation. For them, the neighborhood of a data point $D_i$ is defined in terms of a distance between $\boldsymbol x$. A commonly used metric is the standard Euclidean distance or a scaled variant that reflects the relevance of each feature to the response. We refer to this $\boldsymbol x$-based neighborhood as spatial neighborhood, because GP models are popular spatial statistics models \citep{Cressie1991}, in which the input $\boldsymbol x$ means spatial locations.  Such meaning carries on even for general machine learning purposes \citep{Rasmussen06}.  

In tempGP, due to the expansion in $D_i$, there comes another neighborhood, the temporal neighborhood, defined by the distance between $t$.  The inertia in a physical system means that $\boldsymbol x$ and $y$ will change gradually over time (that is how autocorrelation is resulted), so that the data points in the immediate temporal neighborhood are spatially close too.  When a GP approximation searches for spatial neighbors of a data point, say $\boldsymbol x_0$, it almost surely ends up choosing those data points in the immediate temporal neighborhood of $\boldsymbol x_0$.  Using such a subset for fitting a GP model suffers the same temporal overfitting problem as using the raw data.

TempGP seeks to scramble the temporal neighborhood through thinning, i.e., putting temporally distant data points into the same data block.  Once the original data is scrambled, a new spatial neighborhood can be formed within each block. One may argue that the quality of the new spatial neighborhood is worse than the spatial neighborhood in the raw data and will thus lead to a poor model approximation. We found this argument not necessarily true.  When each block has a sizeable number of data points (say, over $1,000$), it is not difficult to find some data points that are far from $\boldsymbol x_0$ in time but still close to $\boldsymbol x_0$ in value. 

This understanding underpins the possibility of tailoring the existing GP approximations to each block of temporally thinned data points for building a GP approximation with the new data structure. The actual implementations of this idea using twinGP, laGP, scaled Vecchia are different, due to the need to produce a fast enough method with competitive performance in model accuracy. The following subsections present the thinned twinGP, thinned laGP, and thinned SV, respectively.

\subsection{Thinned twinGP}\label{section3.1}
TwinGP blends a sparse global GP with local corrections by modeling the covariance as:
\begin{equation}
k_{\mathrm{total}}(\boldsymbol x, \boldsymbol x') 
\;=\;
(1 - \lambda)\,k_g(\boldsymbol x, \boldsymbol x') 
\;+\;
\lambda\,k_\ell(\boldsymbol x, \boldsymbol x'),
\label{eq:twinGP}
\end{equation}
where \(k_g\) is a long‑range kernel trained on a small, globally selected support set, \(k_\ell\) is a short‑range (compactly supported) kernel used only among spatial nearest neighbors, and \(\lambda\in[0,1]\) balances the broad, smooth trends against the fine, local details. The hyperparameters of the global kernel are learned by approximation of the likelihood at the support points, whereas the local kernel’s radius is set geometrically to ensure that their collective coverage does not leave out any spaces uncovered in the input domain.  The weight \(\lambda\) is chosen by minimizing the prediction error on a hold‑out validation set. Since the global hyperparameters are estimated using a small number of support points and the local kernel is estimated using a small local subset, the method is impressively fast.

Assume that we have already completed the data thinning and grouped the training data into $T$ blocks and suppose we mean to make prediction at a test location \( \boldsymbol x^{*} \).  To apply twinGP, an intuitive approach is to train a twinGP on each of the data blocks and then use each trained GP approximation models to make a prediction, i.e., \(\left\{ \hat{f}_{1}^{*}, \hat{f}_{2}^{*}, \ldots, \hat{f}_{T}^{*} \right\}\). We then average these predictions to obtain a final point estimator for the test locations, i.e.,
\begin{equation}
\bar{f}^* = \frac{1}{T} \sum_{z=1}^T \hat{f}_z^*.
\end{equation}
For each of the GP models, let us denote its predictive standard deviation as $\hat{\sigma}^*_z$.  Then the predictive standard deviation for the averaged prediction is
\begin{equation}
\bar{\sigma}^* = \sqrt{\frac{\sum_{z=1}^T  (\hat{\sigma}^*_z)^2}{T}+\frac{\sum_{z=1}^T  (\hat{f}^*_z-\bar{f}^*)^2}{T}}.
\end{equation}

\begin{figure}[h!] 
\centering
\includegraphics[width=\textwidth, page=1]{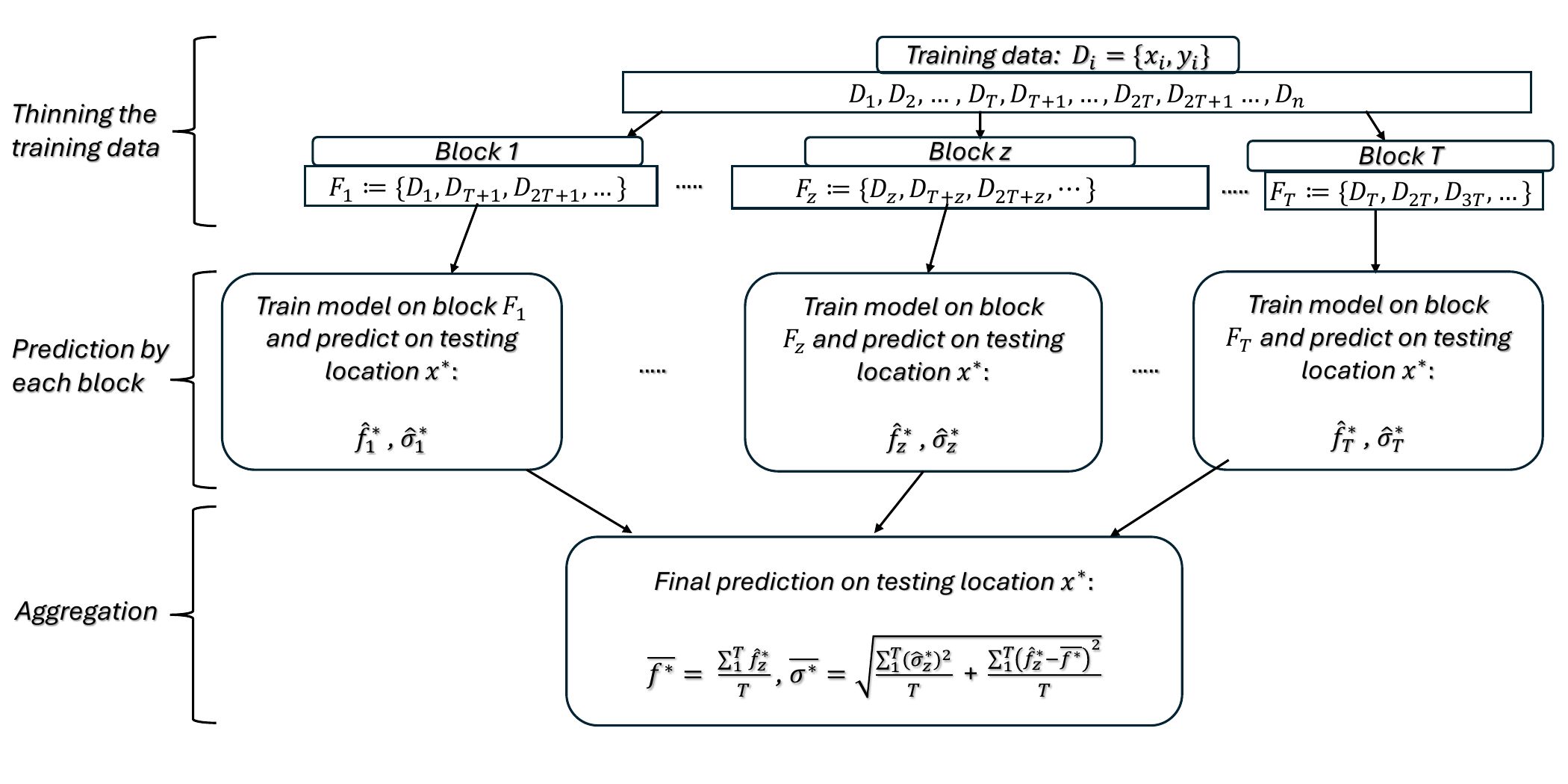}
\caption{Applying twinGP to thinned data via model averaging.}
\label{fig:2step_local_methods1}
\end{figure}

Figure~\ref{fig:2step_local_methods1} illustrates this approach. The main drawback of this approach is its runtime, as it requires training $T$ separate models and could slow down computation. But since twinGP is highly efficient, training $T$ models does not appear a problem.  A thinned twinGP based on the above strategy is still fast overall. 

We would like to note that the idea behind thinned twinGP shares conceptual similarities with divide-and-conquer GP models, such as those in \citet{Ng2014}, where predictions from GPs trained on partitioned data are combined using inverse-variance weighting. We experimented with inverse-variance weighting in thinned twinGP but did not find that doing so presents any advantage. Another alternative is the robust Bayesian committee machine introduced by \citet{Deisenroth2015}, which can improve prediction when there is a clear rationale, such as unequal training set sizes or known deficiencies in certain experts (i.e., blocks in our setting). In such cases, applying penalties or correction weights may be beneficial. However, learning these corrections from data via validation introduces additional complexity and conflicts with our goal of maintaining a fast and interpretable method.

\subsection{Thinned laGP}\label{section3.2}
LaGP avoids the expensive inversion of a large matrix by fitting a small Gaussian process around each test point. For a given test point \(\boldsymbol x_*\), it starts with a small set of the nearest neighbors and then adds one point at a time—each chosen to most reduce the predictive variance at \(\boldsymbol x_*\). These additions are handled via fast rank‑one Cholesky updates rather than refitting the entire model. Once the local subset is fixed, laGP re‑estimates its hyperparameters (length‑scale and noise variance) on just those points. 

Because each local GP (for each test location) tunes its own parameters, laGP naturally adapts to changes across the input space. The downside is that the iterative local point selection and parameter fitting per test location slows laGP down considerably than the alternative GP approximations. Because of this shortcoming, if we apply laGP to the thinned data blocks as we do with twinGP, the resulting thinned laGP would be very slow, sometimes slower than the original tempGP, so defying the purpose of having a fast approximation. 

As such, the approach for using laGP on the thinned data is to use a single thinned block to match with the test location and only use that block for training and prediction for this specific test location. To find which thinned block to be matched with $ \boldsymbol x^*$, it is reasonable to use the nearest neighborhood strategy, which is to find the training location $ \boldsymbol  x_i$ that has the smallest Euclidean distance to $ \boldsymbol  x^*$.  Then locate the data block that contains $\boldsymbol  x_i$; suppose it is $F_c$.  Then $F_c$ is the data block on which we train a laGP and then make prediction for $ \boldsymbol x^*$. This approach does speed up but it is not difficult to see its limitation---it uses only a subset of the data.  The value of $T$ in our applications ranges from 6 to 30.  This means that the single data block approach uses 3--15\% of the original data for training and prediction. As a result, the model accuracy performance of such approach is worse than the averaging approach as in thinned twinGP. We only use this single block approach for laGP because we do not have a better alternative.

\subsection{Thinned scaled Vecchia method}\label{section3.3}
Applying SV to the thinned data using the averaging approach as in the thinned twinGP suffers from a similar slowdown as the approach does to laGP (although not as severe), whereas using the single block approach does not render the full model performance potential as SV could achieve.  On the other hand, the scaled Vecchia procedure allows some additional tailoring to strike a better balance between speed and performance.

The basic idea of Vecchia approximation is to factorize the likelihood into a product of conditional distributions~\citep{Vecchia1988}, in which each data point is treated as conditionally independent of the spatially distant data points, given a subset of spatially nearby data points known as the \emph{conditioning set}.  If the conditioning set is of size $m$, then GP computation can be reduced from \(O(n^3)\) to \(O(n \cdot m^3)\).  The specific version of Vecchia approximation explored in this study is the scaled Vecchia method~\citep{katzfuss2022}, which is so named due to its use of anisotropic scaling. The scaled spatial distance between two locations, $\boldsymbol x_i$ and $\boldsymbol x_j$, is defined as:
\begin{equation}
d(D_i, D_j) := \| \tilde{\boldsymbol{x}}_i - \tilde{\boldsymbol{x}}_j \|,
\end{equation}
where \(\tilde{\boldsymbol{x}} = \left(\frac{x_1}{\ell_1}, \ldots, \frac{x_d}{\ell_d}\right)\) represents the scaled inputs and \(\boldsymbol{\ell} = (\ell_1, \ldots, \ell_d)\)
 are length-scale parameters iteratively updated through some iterations of Fisher-scoring algorithm \citep{katzfuss2022}.  This scaling enhances model efficiency by deactivating irrelevant dimensions while preserving the influence of key variables. The SV approximation is demonstrated to be one of the most computationally efficient methods for GP modeling \citep{katzfuss2022,Kang2021}. 

\begin{figure}[ht]
\centering
\includegraphics[width=\textwidth]{"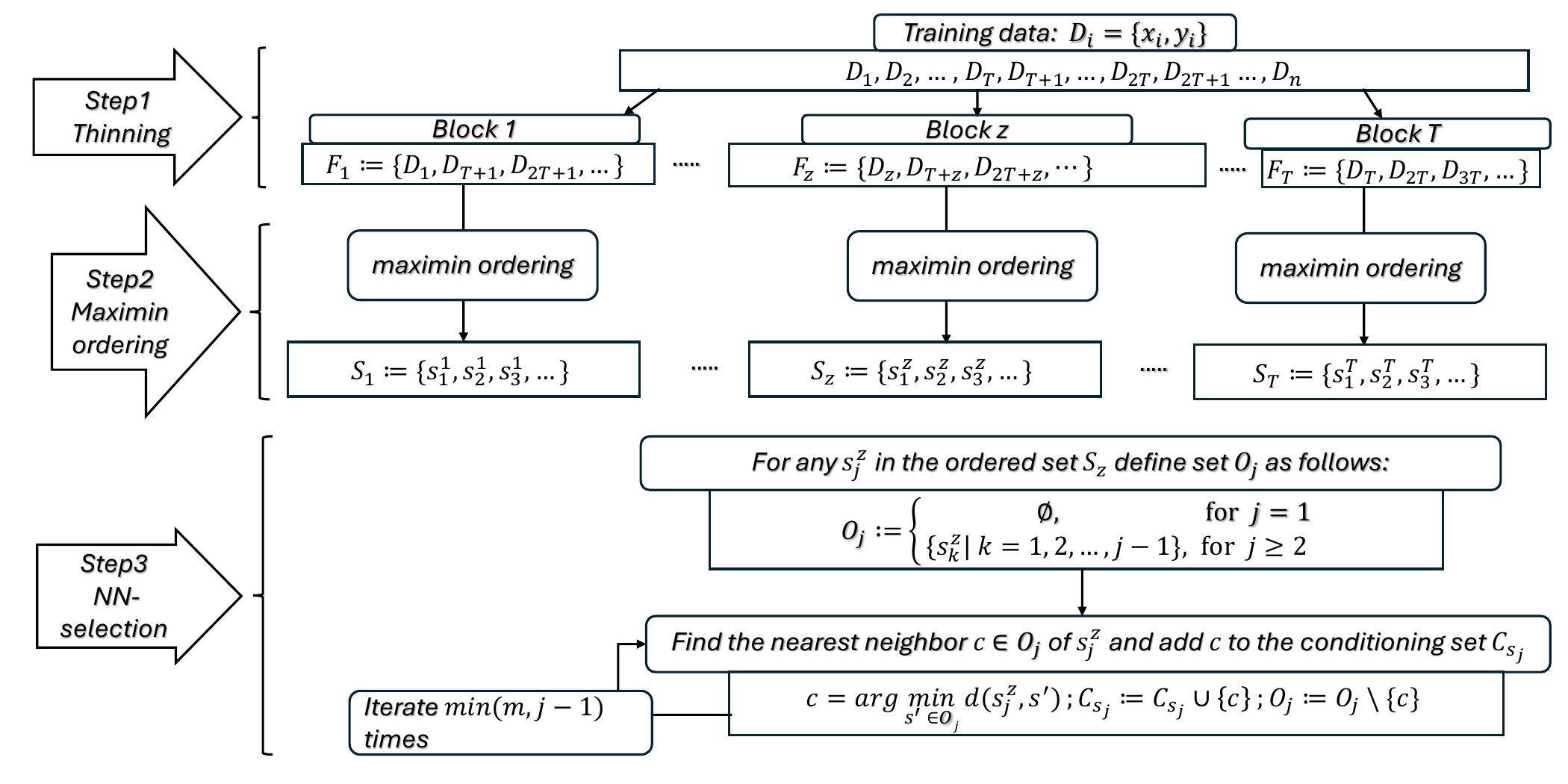"}
\caption{The mechanism for forming the conditioning set in the thinned scaled Vecchia}
\label{fig:3step}
\end{figure}

One of the most important questions in the Vecchia approximation is how to form the conditioning set; for that, SV enhances spatial spread by applying maximin ordering before selecting data points from a spatial nearest neighborhood \citep{katzfuss2020}. Maximin ordering begins by randomly selecting the first data point, and then it chooses each subsequent data point in a manner that maximizes the minimum spatial distance to the previously selected points. In a sense, we still apply SV to each block of the thinned data, but the new data structure requires adjustment to the operations of maximin and nearest neighbor selection. Figure~\ref{fig:3step} illustrates the three-step procedure for thinned SV.  

The first step of thinning is inherited from tempGP, which is the same as in thinned laGP and thinned twinGP.  The following explains the other two steps.
\begin{enumerate}
    \item \textbf{Maximin ordering:} Within each block $j$, we apply the efficient maximin ordering algorithm proposed by \citet{Schafer2021}, yielding the ordered set $S_j$.

    \item \textbf{Nearest neighbors selection:} Denote by $C_{s_j}$ the conditioning set for data point $s^z_j$ the $j$th data point in the ordered block $z$.  The data points in $C_{s_j}$ are selected among the earlier points according to the ordered sequence in $S_z$. We introduce $O_j$ for $s^z_j$, which includes only the data points preceding $s^z_j$ in the ordered set $S_z$. By definition, $O_1$ is empty, $O_2$ includes only $s^z_1$, $O_3$ includes both $s^z_1$ and $s^z_2$, and so on. Then, for $s^z_j$, we choose its $m$ nearest neighbors from $O_j$ or choose the number of the nearest neighbors until $O_j$ is exhausted if the number of data points in $O_j$ is fewer than $m$, and use these nearest neighbors to form $C_{s_j}$.  For data points $s^z_1$ to $s^z_j$, where $j \leq m$, the conditioning set will have the same size as that of $O_j$, which is $j-1$. For the latter data points, the size of the conditioning set is $m$.
    \end{enumerate}

Let us reflect on the difference that the modified data selection procedure makes.  In the original SV, the ordered set of data points are likely conditioned on a diverse set of locations in the beginning, due to the maximin ordering action. However, as one progresses in creating the ordered set, the available choices for conditioning expand significantly. The last point in the ordered set can in theory condition on the entire original dataset. When that happens, the nearest neighbors selected based on the spatial neighborhood criterion are almost surely temporally close, too. As a result, the conditioning sets formed will predominantly include temporally correlated neighbors, and consequently, using such conditioning sets cannot help avoid the temporal overfitting problem in model estimation and prediction. In contrast, when the data selection is restricted to each block, one can be sure that the data points within each conditioning set are not autocorrelated. In the same time, by the nature of the nearest neighbor selection (which is based on the spatial neighborhood criterion), similar enough data points are still chosen to form the conditioning sets. The modified procedure shown in Figure~\ref{fig:3step} carries out the trick to form the conditioning sets by finding similar enough (in $\boldsymbol x$) data points but not from a temporal neighborhood (i.e., make sure the chosen data points are distant apart in $t$), so that the resulting method still enjoys the benefit of SV but can avoid temporal overfitting when the original data are autocorrelated.

\subsection{Estimation of \texorpdfstring{$g(t)$}{g(t)}}

The application of twinGP, laGP, and SV to thinned data in Sections~\ref{section3.1} to~\ref{section3.3} is just for speeding up the estimation of $f(\boldsymbol x)$ in Equation (\ref{Model2}).  The estimation of \( g(t) \) will follow the treatment in the original tempGP.  This is to say, same as the two-step procedure in \cite{Prakash2023}, first estimate $f(\cdot)$ but using the fast approximation now, and then use the residual, $y-\hat{f}(\cdot)$, to estimate $g(\cdot)$.  The GP model used by \cite{Prakash2023} for modeling $g(\cdot)$ is a GP with a zero mean and a standard covariance function. \cite{Prakash2023} used a Matérn covariance function with the shape parameter, $\nu = 1.5$, but other commonly used covariance functions work too, such as the squared exponential function. By the nature of $g(\cdot)$, we know it is a localized function. The autocorrelation for the current time instance should not extend beyond $\pm T$ from the current time index. Therefore, when estimating $g(\cdot)$, we simply use the temporal neighborhood data from $[t-T, t+T]$ for any time point $t$.  As $T$ is a small value (usually in tens and definitely smaller than hundreds), the estimation of $g(\cdot)$, even using the standard GP estimation, does not take much time. So no fast GP approximation is needed.

\subsection{Prediction}
\label{Prediction}
The two terms in Equation~(\ref{Model2}), $f(\cdot)$ and $g(\cdot)$, are treated differently. 

\vspace{6 pt}
\noindent \textbf{Prediction of $f(\cdot)$}:
\begin{itemize}
\item For thinned twinGP, the model averaging described in Section~\ref{section3.1} is used to making a prediction of $f(\cdot)$ at $\boldsymbol x^*$. 
\item For thinned laGP, the single block approach described in Section~\ref{section3.2} is used to making a prediction of $f(\cdot)$ at $\boldsymbol x^*$
\item For thinned SV, the prediction $f(\cdot)$ at $\boldsymbol x^*$ is to use another conditioning set of the size of $m_p$.  Let us denote this conditioning set by $C^*$. To create $C^*$, we again use maximin ordering and nearest neighbor selection to choose $m_p$ data points for prediction at $\boldsymbol x^*$. 
    While similar in spirit, the conditioning set for prediction is not the same as the previous conditioning set for estimation. There are two main differences here:
    \begin{itemize}
    \item[-] The candidate set of size $m_p$ for prediction includes the test locations for which the predictions are already made, whereas the conditioning set of size $m$ for training includes only the training locations. Generally speaking, $m_p > m$.
    \item[-] The maximin ordering is conducted on the original whole set of training data and test data without data thinning and blocking.
    \end{itemize}
    Specifically, one first performs maximin ordering separately for the whole training and test datasets, resulting in the ordered training set and the ordered test set. For the prediction on the first location in the ordered test set, the selection of the nearest neighbors is from the ordered training set. Once made the prediction, add the first location in the ordered test set to the ordered training set to form an augmented set, in which the nearest neighbor selection will be conducted for the next prediction.  This procedure repeats itself until all predictions are made.
\end{itemize}

\textbf{Prediction of $g(\cdot)$}: The prediction of $g(\cdot)$ is done subsequently by using the residuals, $y-\hat{f}(\cdot)$. Here, the prediction on $g(\cdot)$ can be done rather quickly because only the data temporally close to the target time point $t^*$ are used to make the prediction.  As we explain earlier,  $g(\cdot)$ captures the autocorrelation in data, but for many engineering systems, autocorrelation dies down quickly, so that $g(\cdot)$ has a compact support.

\textbf{Overall prediction}: The final prediction is the summation of $\hat{f}(\boldsymbol x^*)$ and $\hat{g}(t^*)$. If the original dataset has very little temporal correlation among its data points, or for a prediction location $\boldsymbol x^*$ uncorrelated with any existing training locations, $\hat{g}(t^*)$ will return a near zero value, meaning that the effective prediction reduces to $\hat{f}(\boldsymbol x^*)$.

\section{Numerical experiments}
\label{Numerical Experiments}
We conduct numerical experiments to compare the performance of the three GP approximations and their thinned counterparts.  For SV, we implement two versions---one uses $\boldsymbol x$ as input while ignoring the time information, whereas the other cascades $\boldsymbol x$ and $t$ into a combined input. We denote them by SV($\boldsymbol x$) and SV($\boldsymbol x, t$), respectively. The reason of including SV($\boldsymbol x$,t) is because one may ask whether including $t$ in the inputs while using the scaled Vecchia approximation naturally takes care of the temporal overfitting problem, considering $t$ is part of the inputs in the tempGP model.  The short answer is that doing so helps under some circumstances but not always.  Including SV($\boldsymbol x, t$) in the comparison provides the numerical evidence.

One tunable parameters common to the three approximations is the thinning parameter, $T$, because all approximations are applied to the data blocks after thinning. The parameter $T$ is chosen according to Equation~(\ref{thinning_x&y}).

Thinned SV has two additional tunable parameters: the size of the conditioning set in training, $m$, the size of the conditioning set in prediction, $m_p$, and the thinning number, $T$.  The choice of these parameters determines the balance between prediction accuracy and computational efficiency.  The sizes of the conditioning sets for estimation, \( m \), and for prediction, \( m_p \), play a vital role. While smaller \( m \) values (recommended between 10 and 30) suffice for estimation purposes, enhancing model efficiency without significantly sacrificing accuracy, larger \( m_p \) values are preferable for prediction to improve accuracy. However, \( m_p \) values larger than 400 tend to increase runtime disproportionately, suggesting a practical upper limit for this parameter. For all numerical experiments, we use $m=30$ and $m_p=140$ for thinned SV and both versions of SV because the same parameter values have been used in the previous SV analysis~\citep{katzfuss2022}. 

LaGP has an additional tunable parameter, the neighborhood size corresponding to \(m\) in SV, so we set it at \(30\).

TwinGP has three additional tunable parameters: the global‐set size, the local‐set size, and the hold‑out validation‑set size. In the standard twinGP implementation, these are automatically selected based on the dimension of the training set. But if we leave these three parameters still automatically selected by the existing twinGP method on the thinned data blocks, these parameters will all become smaller because the data size of each block is smaller than that of the original dataset.  Using the smaller size parameters in thinned twinGP leads to bad model performance.  Our fix for this problem is to apply twinGP on the original dataset once and record the three size parameters it selected.  Then use the same parameters when a twinGP is fit to the data in each thinned block.  This treatment greatly improves the performance of thinned twinGP without affecting its runtime much.

The covariance function for laGP and the global term of twinGP is the Gaussian kernel, while twinGP’s local kernel is a compactly supported radial‐basis function. For the Vecchia‐based models, we employ the Matérn covariance with smoothness \(\nu=1.5\) to match the settings of tempGP. 

All experiments were performed using five Intel CPU cores on the RHL7 Phoenix cluster, part of the leading-edge services provided by the Partnership for an Advanced Computing Environment (PACE) High-Performance Computing (HPC) facility at the Georgia Institute of Technology. \cite{PACE2017} offers a managed, scalable, and research-optimized computational environment, ensuring consistent performance and reproducibility across experiments.

In this numerical study, we use four datasets: the robotic arm simulation data~\citep{Surjanovic13}, the satellite drag coefficient data~\citep{Gramacy2015}, and two real datasets, known as DSWE Dataset 5 and DSWE Dataset 6, respectively, which are the companion datasets for the book, \emph{Data Science for Wind Energy} \citep[DSWE]{Ding2019}. On the satellite drag coefficient data and DSWE Dataset 5, a randomized cross-validation is used to evaluate the performance model.  On the robotic arm simulation data, GP models are trained on one set of data and tested on an entirely different set of newly simulated data.  On the DSWE Dataset 6, GP models are trained on a set of historical data of one year and tested on the set of data of a different year. More details about these datasets are given in the following subsections. 

For the robotic arm simulation data and the DSWE Dataset 6, $t$ in their test sets is far distant from the time indices in the training sets, and thus the $g(\cdot)$ term in Equation (\ref{Model2}) becomes effectively zero. What is being tested is just the $f(\cdot)$ term. Under such conditions, SV($\boldsymbol x, t$) implementation is not expected to perform well due to extreme extrapolation in time dimension. For the satellite drag coefficient data and the DSWE Dataset 5, the $g(\cdot)$ term is not necessarily zero and what is being tested is the combined effect of $f(\cdot)$ and $g(\cdot)$.  We want to note that we did not eliminate $g(\cdot)$ or manually set it to zero, but just let the models decide on it based on the data and circumstances. 

We use two metrics to evaluate the model performance: the Root Mean Squared Error (RMSE) and Negative Log Predictive Density (NLPD). RMSE evaluates the precision of the model's point prediction, whereas NLPD assesses how well the predicted posterior distribution aligns with the test data. Smaller RMSE and NLPD values indicate better model performance. 

For a test set \(\{\boldsymbol x^*_i, y^*_i\}\) consisting of \(n_\text{test}\) test locations, RMSE and NLPD are defined, respectively, as follows:
\begin{equation}
    \label{rmse}
    \text{RMSE} = \sqrt{\frac{1}{n_\text{test}} \sum_{i=1}^{n_\text{test}} \left(y^*_i - \hat{\mu}(\boldsymbol x^*_i)\right)^2},
\end{equation}

\begin{equation}
   \label{nlpd}
    \text{NLPD} = \frac{1}{2n_\text{test}} \sum_{i=1}^{n_\text{test}} \left[ \frac{\left(y^*_i - \hat{\mu}(\boldsymbol x^*_i)\right)^2}{\hat{\sigma}^2(\boldsymbol x^*_i)} + \log\left(2\pi \hat{\sigma}^2(\boldsymbol x^*_i)\right) \right],
\end{equation}
where $\hat{\mu}(\boldsymbol x^*_i)$ and $\hat{\sigma}^2(\boldsymbol x^*_i)$ are the mean and variance of the prediction at $\boldsymbol x^*$, respectively.

For thinned laGP, thinned twinGP, and thinnned SV, because of the existence of \(g(\cdot)\) in the model structure, determining the predictive variance becomes tricky.  In the end, we decide to approximate the predictive variance by adding \(\text{var}(f(\cdot))\) and \(\text{var}(g(\cdot))\). Recall that we fit \(g(\cdot)\) to the residuals after taking out the effect of \(f(\cdot)\).  We find that the magnitude of \(\text{var}(g(\cdot))\) is about 1\% of that of \(\text{var}(f(\cdot))\). Ignoring the covariance between \(f(\cdot)\) and \(g(\cdot)\) does not fundamentally change the overall picture. Please note that the approximation is only applicable to the satellite drag coefficient data and the DSWE Dataset 5. For the robotic arm simulation data and the DSWE Dataset 6, since \(g(\cdot)\) is effectively zero, the predictive variance on these two datasets are not approximated. As a last note, tempGP is only included in the RMSE comparison not in NLPD because tempGP does not produce predictive variances as its model output.

\subsection{Simulation on robot arm function}

A popular simulation dataset used in the previous GP computation comparisons is the robot arm simulation from the Virtual Library of Simulation Experiments \citep{Surjanovic13}. The mathematical representation of the robotic arm calculates the position of the end-effector based on joint angles and segment lengths. The model is defined through several key equations:

\begin{itemize}
\item Given the random variables \(\boldsymbol x = [\Theta_1, \Theta_2, \Theta_3, \Theta_4, L_1, L_2, L_3, L_4]\) as input, we need to compute the distance \(y\) from the end of the robotic arm to the origin.
First, compute the cumulative sum of angles:\begin{equation}
\xi_i = \sum_{j=1}^{i} \Theta_j \quad \text{for } i = 1, 2, 3, 4\end{equation}

\item Next, compute the components \(u\) and \(v\):\begin{equation}
u = \sum_{i=1}^{4} L_i \cos(\xi_i)\end{equation}
\begin{equation}
v = \sum_{i=1}^{4} L_i \sin(\xi_i)\end{equation}

\item Finally, the distance \(y\) from the end of the arm to the origin is given by:\begin{equation}
y = \sqrt{u^2 + v^2}\end{equation}

\end{itemize}
These formulations explains how the end-effector's position is derived from the joint angles and segment lengths, demonstrating the operational reach of the robotic arm in a 2D plane.  

\begin{figure}[ht]
\centering
\includegraphics[width=.8\textwidth]{"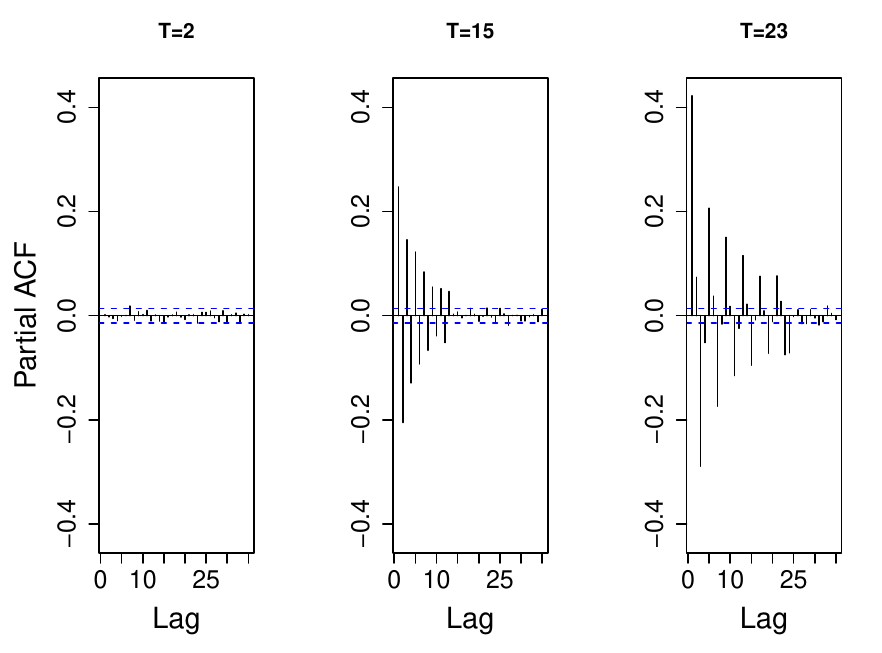"}
\caption{The left plot displays the PACF for the LHS dataset, which shows no significant autocorrelation. The middle and right plots use different lag values, $\mathbf{M=13}$ and 
$\mathbf{M=24}$, respectively. The corresponding thinning numbers are 2, 15, and 23. In both the middle and right plots, partial autocorrelation persists over moderate and longer lags, indicating stronger temporal dependence.}
\label{fig:acfplots}
\end{figure}

In this study, we create 3 scenarios. In the first scenario, we sample data using random Latin hypercube design. For the next two scenarios, we modify this simulation by injecting moderate and high temporal correlation into its data. Incorporating temporal correlation in the context of robot-arms makes physical sense because the joint angles and segment lengths cannot change instantaneously; they should exhibit smooth, continuous transitions over time, reflecting the physical constraints of mechanical movement.  To inject autocorrelation, an autoregressive structure is now used to simulate the joint angles $\Theta_{i,t}$ and segment lengths $L_{i,t}$:

\begin{align}
\Theta_{i,t}' &= \sum_{k=1}^M \phi_{i,k} \Theta_{i,t-k} + \epsilon_{i,t},  \\
L_{i,t}' &= \sum_{k=1}^M \phi_{i,k} L_{i,t-k} + \epsilon_{i,t},\label{eq:l_ar_model}
\end{align}
where $\phi$ is a set of coefficients leading in stationary autocorrelated process for covariates and $\epsilon$ is random noise and  $M$ is the parameter used to control the degree of autocorrelation. The temporal correlation is further transmitted to $y_t$ via $u_t$ and $v_t$:

\begin{align}
u_t &= \sum_{i=1}^{4} L_{i,t}' \cos\left( \sum_{j=1}^{i} \Theta_{j,t}' \right), \\
v_t &= \sum_{i=1}^{4} L_{i,t}' \sin\left( \sum_{j=1}^{i} \Theta_{j,t}' \right), \\
y_t &= \sqrt{u_t^2 + v_t^2} + \sum_{k=1}^M \psi_{k} \epsilon_{t-k},  \label{eq:y_t}
\end{align}
where $\psi$ is a set of different coefficients leading to different autocorrelated stationary processes for autocorrelated noise. One way to measure autocorrelation in real data is the thinning parameter $T$, i.e., how large $T$ is such that the thinned data in a block is virtually uncorrelated. Analyzing the datasets used in this study reveals that $T$ is in the range of 6 to 26 for $M=13$ and $M=24$. We tune $M$ to set the degree of autocorrelation to match those in the real data.  
Figure \ref{fig:acfplots} presents the partial autocorrelation function for three different levels of autocorrelation on $L_1$, one of the input variables. 

The simulated data has eight input variables, four $\Theta$'s and four $L$'s, so that \(\boldsymbol{x}\) is of the dimension of $8\times 1$.  We simulate 20,000 data points, i.e., $n=20,000$, for training.  For testing, we simulate an entirely new set of data of 10,000 observations. Since this dataset does not have a natural time index, $t$ here is simply sequential number of all $30,000$ data points. 

To account for variability and evaluate model robustness, we repeat each simulation scenario over 10 independent replications. In the first scenario (with no autocorrelation), each replication begins by generating a new training dataset using an independently drawn Latin hypercube design (LHS). This introduces randomness through the sampling of covariate values. In the second and third scenarios (with moderate and high autocorrelation), data are generated using autoregressive (AR) processes. In these cases, randomness arises from two sources: (i) the initial $M$ values of each AR process, which are randomly drawn from a standard normal distribution, and (ii) the innovation noise terms $\epsilon_{i,t}$ and $\epsilon_t$, also sampled from a Gaussian distribution. Importantly, both the training and testing datasets are regenerated for each replication, ensuring that performance metrics reflect variability across different realizations of the underlying stochastic process. Table~\ref{tab:robot_perf} presents model performance comparison in terms of RMSE and NPLD. We observe the following:
\begin{itemize}
\item In presence of moderate and strong autocorrelation, the thinned version always improves their counterpart in terms of both RMSE and NLPD for SV and twinGP.  The improvement effect for thinned SV over SV (16\%) is more pronounced than thinned twinGP over twinGP (2\%).  When comparing all the methods, thinned SV is the winner in terms of RMSE and thinned twinGP is better in terms of NLPD. The performances of thinned SV and thinned twinGP are close, differing by no more than 2\%.  Thinned laGP is not competitive.

\item In low autocorrelation, thinned laGP and thinned twinGP are worse than their counterparts, but thinned SV could perform close to SV from the RMSE aspect and slightly weaker in terms of NLPD. For this data, SV($\boldsymbol x$, $t$) emerges as the best performer in terms of both RMSE and NLPD. TwinGP and its thinned version performed two and three times worse than SV methods in terms of RMSE.
\item In this simulation data, using the combined input ($\boldsymbol x, t$) does not appear to make much difference in terms of both RMSE and NLPD.
\end{itemize}

\begin{table}[ht]
\centering \footnotesize
\caption{\centering  \footnotesize Performance comparisons for the robot arm simulation data. 
Boldface font indicates the best performance among the GP approximations. The last row is the results of tempGP, which are included as a full GP benchmark.}
\begin{tabular}{|l|c|c|c|c|c|c|c|}
\hline
 &\multicolumn{3}{c|}{Mean  $\pm$  SD RMSE} & \multicolumn{3}{c|}{Mean  $\pm$  SD NPLD}\\
 \hline
 autocorrelation level& low  &  moderate  & high   & low  &  moderate  & high   \\
 \hline
laGP &  0.052 $\pm$ 0.001  & 9.052 $\pm$ 0.207 & 7.764 $\pm$ 0.235 & -1.195 $\pm$ 0.040 & 6.756 $\pm$ 0.189 & 5.900 $\pm$ 0.087 \\
\hline
twinGP &0.028 $\pm$ 0.002 & 6.954 $\pm$ 0.160& 6.104 $\pm$ 0.200& -2.253 $\pm$ 0.060 &3.332 $\pm$ 0.020 &3.199 $\pm$ 0.030
 \\
\hline
SV($\boldsymbol x$) & \textbf{0.013 $\pm$  0.000}& 8.102 $\pm$ 0.163& 7.041 $\pm$ 0.173
& -2.977 $\pm$ 0.013& 4.297 $\pm$ 0.069& 3.314 $\pm$ 0.023\\
\hline
SV($\boldsymbol x$,$t$) &\textbf{0.013 $\pm$  0.000} &8.022 $\pm$ 0.165& 6.943 $\pm$ 0.177&\textbf{ -2.979 $\pm$ 0.012} &4.274 $\pm$ 0.130 &3.293 $\pm$ 0.024\\
\hline
thinned laGP & 0.059 $\pm$ 0.001 & 9.472 $\pm$ 0.284 & 8.294 $\pm$ 0.261
& -1.014 $\pm$ 0.050 &6.581 $\pm$ 0.203 & 5.812 $\pm$ 0.140\\
\hline
thinned twinGP & 0.037 $\pm$ 0.020 & 6.824 $\pm$ 0.120 &5.966 $\pm$ 0.154& -1.997 $\pm$ 0.487 &\textbf{3.315 $\pm$ 0.015} &\textbf{3.174 $\pm$ 0.020} \\
\hline
thinned SV & \textbf{ 0.013 $\pm$ 0.000}& \textbf{6.809 $\pm$ 0.111}& \textbf{5.936 $\pm$ 0.129}& -2.970 $\pm$ 0.021  &3.346 $\pm$ 0.017 &3.197 $\pm$ 0.021 \\
\hline
tempGP &0.019 $\pm$ 0.000 &7.015 $\pm$ 0.180  &5.909
 $\pm$ 0.139  & - & - & - \\
\hline
\end{tabular}
\label{tab:robot_perf}
\end{table}

Table \ref{tab:runtime_robot} presents the runtime (including both training and prediction) for the seven methods averaged over all autocorrelation levels. The original twinGP is much faster than any other methods, which is its noteworthy advantage.  The thinned twinGP slows down considerably, but is still more than two times faster than thinned SV.  All methods except laGP and thinned laGP are fast enough to finish around three minutes or less.  Both laGP and thinned laGP are slower than the other GP approximations.

As a reference, we also run tempGP on the simulated data.  The RMSEs of tempGP are 0.019, 7.015, and 5.909 at the low, medium, and high autocorrelation levels, respectively.  The best thinned version of the GP approximations actually does better than tempGP for low and moderate autocorrelation, but slightly worse for high autocorrelation. The runtime of tempGP is on average 2.3 hours (i.e., 8,234 seconds), significantly slower than thinned twinGP or thinned SV. 

\begin{table}[ht]
\centering
\footnotesize
\caption{\footnotesize Runtime averaged over replications and levels of autocorrelation (in seconds)}
\begin{tabular}{|l|c|c|c|c|c|c|c|c|}
\hline
Method & laGP & twinGP & SV($\boldsymbol x$) & SV($\boldsymbol x, t$) & thinned laGP & thinned twinGP & thinned SV& tempGP\\
\hline
Runtime (sec) & 2,030 & 6 & 93 & 93 & 1,649 & 81 & 188 & 8,234\\
\hline
\end{tabular} \label{tab:runtime_robot}
\end{table}

\subsection{Satellite drag coefficient data}
\cite{Gramacy2015} provided the satellite drag coefficients datasets. We use seven of their datasets, four of them are from \texttt{hst} satellite and labeled as \texttt{hstA}, \texttt{hstA\_05}, \texttt{hstQ}, and \texttt{hstQ\_05}, respectively. The other three are from \texttt{grace} satellite and labeled as \texttt{graceA\_05}, \texttt{graceQ}, and \texttt{graceQ\_05}, respectively. All datasets are of the same size, 8,641 data points, and all of them are time series in that each data point represents measurements taken at regular 10-second intervals, including the total drag coefficient as response and a set of features as inputs. The inputs include \( v_{\text{rel}}, T_{s}, T_{a}, \theta, \phi, \alpha_{n}, \sigma_{t} \).  However, some of these features, like \( T_{s} \), which is the surface temperature, are constant on the whole datasets and thus treated as non-informative predictors.  As such, the input variables for our setting are reduced to three: \( T_{a}, \alpha_{n}, \) and \( v_{\text{rel}} \), standing for atmospheric temperature, normal energy, and velocity, respectively. 

As mentioned earlier, a randomized 5-fold cross-validation is used on the satellite drag coefficients datasets to evaluate the respective RMSE and NLPD for the seven methods. Tables \ref{tab:satel_dataset_RMSE} and \ref{tab:satel_dataset_NLPD} present the model performance comparison with RMSE and NLPD, respectively. For this dataset, thinned twinGP and thinned SV turn out to be the best methods and they have comparable performance with each other.  Interestingly, the original twinGP, when applied to this particular set of data, also delivers an outstanding performance, just slightly worse than thinned SV and thinned twinGP. Like in the robot arm simulation data, thinned SV once again makes a remarkable improvement over SV. LaGP and its thinned counterpart are not competitive. We also observe that on this dataset using cross-validation, including $t$ as part of the input helps SV, although marginally. 

\begin{table}[t]
\centering \footnotesize
\caption{\footnotesize RMSE comparisons across seven datasets (values scaled by $\mathbf{10^{3}}$ for clarity). Boldface font indicates the best performance among the GP approximations.}
\begin{tabular}{|l|c|c|c|c|c|c|c|c|}
\hline
 & \multicolumn{8}{c|}{RMSE} \\
\hline
 & \texttt{hstA} & \texttt{hstA\_05} & \texttt{hstQ} & \texttt{hstQ\_05} & \texttt{graceA\_05} & \texttt{graceQ} & \texttt{graceQ\_05} & Average \\
\hline
laGP & 4.39 & 4.01 & 8.92 & 17.97 & 8.71 & 3.37 & 3.86 & 7.32 \\
\hline
twinGP & 3.44 & 2.94 & 6.31 & \textbf{6.15} & 6.87 & 3.12 & 3.62 & 4.64 \\
\hline
SV(\(\boldsymbol x\)) & 5.21 & 4.61 & 9.06 & 9.14 & 9.72 & 4.49 & 5.24 & 6.78 \\
\hline
SV(\(\boldsymbol x,t\)) & 5.19 & 4.50 & 8.90 & 9.16 & 9.68 & 4.46 & 5.21 & 6.73 \\
\hline
thinned laGP & 5.30 & 4.85 & 11.87 & 23.00 & 10.82 & 3.86 & 4.35 & 9.15 \\
\hline
thinned twinGP & 3.41 & \textbf{2.90} & 6.27 & 6.17 & \textbf{6.71} & \textbf{3.11} & \textbf{3.60} & \textbf{4.60} \\
\hline
thinned SV  & \textbf{3.39} & \textbf{2.90} & \textbf{6.26} & 6.16 & 6.81 & \textbf{3.11} & 3.61 & 4.61 \\
\hline
tempGP & 3.39& 	2.91& 6.29& 6.13& 6.72& 3.10& 3.60 &4.59\\
\hline
\end{tabular}
\label{tab:satel_dataset_RMSE}
\end{table}

\begin{table}[t]
\centering \footnotesize
\caption{\footnotesize NLPD comparisons across seven datasets}
\begin{tabular}{|l|c|c|c|c|c|c|c|c|}
\hline
 & \multicolumn{8}{c|}{NLPD} \\
\hline
 & \texttt{hstA} & \texttt{hstA\_05} & \texttt{hstQ} & \texttt{hstQ\_05} & \texttt{graceA\_05} & \texttt{graceQ} & \texttt{graceQ\_05} & Average \\
\hline
laGP   & -3.87 & -4.01 & -3.30 & -2.81 & -3.28 & -4.08 & -3.93 & -3.61 \\
\hline
twinGP & -4.19 & -4.36 & -3.57 & -3.65 & -3.51 & -4.31 & -4.19 & -3.97 \\
\hline
SV(\(\boldsymbol x\)) & -2.20 & -2.33 & -2.85 & -2.90 & -2.80 & -3.52 & -3.39 & -2.86 \\
\hline
SV(\(\boldsymbol x,t\)) & -2.22 & -2.59 & -2.95 & -2.94 & -2.86 & -3.60 & -3.51 & -2.95 \\
\hline
thinned laGP & -3.71 & -3.83 & -3.06 & -2.45 & -3.09 & -3.98 & -3.84 & -3.42 \\
\hline
thinned twinGP & \textbf{-4.26} & \textbf{-4.42} & -3.63 & \textbf{-3.66} & -3.52 & \textbf{-4.35} & \textbf{-4.20} & -4.00 \\
\hline
thinned SV & \textbf{-4.26} & -4.41 & \textbf{-3.64} & \textbf{-3.66} & \textbf{-3.56} & -4.34 & -4.19 & \textbf{-4.01} \\
\hline
\end{tabular}
\label{tab:satel_dataset_NLPD}
\end{table}

In terms of runtime, all methods except laGP and thinned laGP can be run in less than a minute; see Table \ref{tab:runtime_statel}. For this dataset, tempGP achieves the average RMSE of $4.59\times 10^{-3}$, matching the performance of thinnedSV and thinned twinGP (the last column in Table~\ref{tab:satel_dataset_RMSE}). TempGP's runtime is 48 seconds for this dataset.

\begin{table}[ht]
\centering
\footnotesize
\caption{\footnotesize Runtime averaged on five folds of all the satellite drag coefficient datasets (in seconds)}
\begin{tabular}{|l|c|c|c|c|c|c|c|c|}
\hline
Method & laGP & twinGP & SV($\boldsymbol x$) & SV($\boldsymbol x, t$) & thinned laGP & thinned twinGP & thinned SV & tempGP\\
\hline
Runtime (sec) & 309 & 1 & 10 & 10 & 547 & 15 & 31 & 48\\
\hline
\end{tabular} \label{tab:runtime_statel}
\end{table}

\subsection{Wind power curve estimation}
The wind turbine power curve, which is a crucial link connecting wind speed and various environmental factors to wind power output, is essential in wind energy operations and planning~\citep{Ding2019}. Accurately modeling a power curve is key for making well-informed decisions in areas like forecasting wind energy and evaluating wind turbine efficiency. 

We implemented all methods on two sets of wind turbine datasets that are publicly accessible: DSWE Dataset 5 and DSWE Dataset 6.  As a brief recap of the two datasets, Dataset 5 contains data records from six wind turbines (WT1 through WT6), whereas Dataset 6 contains data records from four wind turbines (WT1 through WT4). There are both onshore and offshore turbines among the ten turbines.  The inputs for each turbine are the environmental variables, including wind speed and direction, among others.  Depending on the turbines, the number of input variables ranges from five to seven.  The response is the active power produced and measured on a respective turbine.  The original physical unit of power is in megawatts (MW), but this power data in the open datasets has been normalized, with the maximum power being one, and all other power values scaled down proportionally. 

\begin{table}[t]
\centering \footnotesize
\caption{\footnotesize RMSE comparisons for DSWE Dataset 5. Boldface font indicates the best performance among the GP approximations.}
\begin{tabular}{|l|c|c|c|c|c|c|c|}
\hline
 &\multicolumn{7}{c|}{RMSE} \\
 \hline
 & WT1 & WT2 & WT3  & WT4 & WT5 & WT6 & Average \\
 \hline
laGP &7.87 &8.36 &6.85 &10.75 &9.35 &9.54 &8.79 \\
\hline
twinGP &8.78 &9.14 &7.37 &11.59 &9.19 &9.66 &9.29\\
\hline
SV($\boldsymbol x$) &7.75 &8.98 &6.45 &10.82 &9.20 &8.72 &8.65\\
\hline
SV($\boldsymbol x$,$t$) & 7.02&8.61 &\textbf{5.33} &9.66 &8.16 &7.80 &7.76\\
\hline
thinned laGP &6.32 &8.07 &6.09 &9.03 &8.19 &8.52 &7.70\\
\hline
thinned twinGP &6.32 &7.05 &5.38 &7.86 &6.66 &6.63 &6.65\\
\hline
thinned SV &\textbf{5.75} &\textbf{6.72} &5.59 &\textbf{7.65} &\textbf{6.58} &\textbf{6.58} &\textbf{6.48} \\
\hline
tempGP &6.09	
 & 7.09& 4.97& 7.85& 6.56	& 6.56 &6.52 \\
\hline
\end{tabular}
\label{tab:dataset5_RMSE}
\end{table}

Dataset 5 spans one year. When using this dataset, we use a five-fold cross-validation to compute RMSE and NLPD. Dataset 6 includes data for more than two years.  When using Dataset 6, we use the first year data as training data and using the second year data as a separate test dataset. As such, using Dataset 6 allows us to test the methods when the test data are from a time period entirely different from the training period.

Both Dataset 5 and Dataset 6 are from the SCADA (Supervisory Control And Data Acquisition) system of the wind turbines.  A standard data management in the wind industry is that the SCADA system provides the average measurements over a ten-minute interval.  What this means is that the data records in Datasets 5 and 6 are every 10 minutes.  If there is no missing data, then one year's worth of SCADA data is 52,560. The actual amount in Datasets 5 and 6 is fewer than 52,560.  The amount of data associated with each turbine varies, but it is generally around 45,000.

On Dataset 5, for which we conduct 5-fold randomized cross-validation, the thinned version is uniformly better than the original GP approximations. On average, the thinned version has an RMSE 12--28\% smaller.  Thinned SV appears overall the best, followed by thinned twinGP. The difference between the top two performer is 2--3\% in RSME and NLPD.  When doing cross validations, SV($\boldsymbol x, t$) does help reduce both RMSE and NLPD but the effect of improvement is noticeably smaller than the improvement made by the thinned version of the GP approximations. 

\begin{table}[t]
\centering \footnotesize
\caption{\footnotesize NPLD comparisons for DSWE Dataset 5}
\begin{tabular}{|l|c|c|c|c|c|c|c|}
\hline
 &\multicolumn{7}{c|}{NPLD} \\
 \hline
 & WT1 & WT2 & WT3  & WT4 & WT5 & WT6 & Average \\
 \hline
laGP &3.25 &3.41 &\textbf{3.14} &3.68 &3.41 &3.48 &3.47 \\
\hline
twinGP &3.59 &3.62 &3.39 &3.87 &3.62 &3.67 &3.63\\
\hline
SV($\boldsymbol x$) & 3.62&3.74 & 3.38&3.92 &3.79 &3.72 &3.70\\
\hline
SV($\boldsymbol x$,$t$) &3.51 &3.68 &3.20 &3.80 &3.61 & 3.59&3.57\\
\hline
thinned laGP &3.44 &3.55 & 3.18 &3.69 &3.65 &3.71 &3.54\\
\hline
thinned twinGP &3.44 &3.49 &3.34 &3.68 &3.41 & 3.48&3.47\\
\hline
thinned SV &\textbf{3.20} &\textbf{3.34} &3.29 &\textbf{3.51} &\textbf{3.38} &\textbf{3.44} &\textbf{3.36} \\
\hline
\end{tabular}
\label{tab:dataset5_NPLD}
\end{table}

On Dataset 6, thinned SV is 4\% better than thinned twinGP in terms of RMSE, but thinned twinGP is 4\% better than thinned SV in terms of NLPD. On this dataset, in which we use the 2nd year data to evaluate a model built on the 1st year data, SV($\boldsymbol x,t$) apparently does not help. Its RMSE is several-fold larger than the rest of the methods. NLPD of SV($\boldsymbol x,t$) is 29\% larger than that of  SV($\boldsymbol x$). This occurs because the test data’s time index falls in a different year, resulting in both a large mean prediction error and a high predictive variance.

Table~\ref{tab:runtime_wind} presents the runtime on both Dataset 5 and Dataset 6 for all methods.  TwinGP is still the fastest, which makes the thinned twinGP fast too.  Thinned SV takes around 4--9 minutes.  Please note this is the combined runtime for training and prediction(average runtime per fold for dataset 5).  For training alone, it is half of this time, i.e., around 2--5 minutes.  For Dataset 5, it is the 

For reference, tempGP requires approximately 3 hours to complete training and prediction on the full size of Dataset 6, and around 1.5 hours for training on four folds and prediction on the remaining fold of Dataset 5. The runtime of tempGP on high-performance computers is significantly faster. In contrast, running a single Dataset 6 experiment on a personal laptop takes 21 hours, whereas approximation methods like thinnedSV are only slightly slower, taking less than a minute longer. The average RMSEs of tempGP when applied to Dataset 5 and Dataset 6 are 6.52 and 3.56, respectively, and in both cases bigger than that of the best thinned approximation, i.e., the thinned SV's average RMSEs (6.48 and 3.28, respectively). 

\begin{table}[ht]
\centering \footnotesize
\caption{\footnotesize Performance comparisons for DSWE Dataset 6. Boldface font indicates the best performance among the GP approximations.}
\begin{tabular}{|l|c|c|c|c|c|c|c|c|c|c|c|}
\hline
 &\multicolumn{5}{c|}{RMSE} & \multicolumn{5}{c|}{NPLD}\\
 \hline
 & WT1 & WT2 & WT3  & WT4 & Average & WT1 & WT2 & WT3  & WT4 & Average\\
 \hline
laGP &4.12&4.31 &4.50 &4.40 &4.33 & 3.14&3.23 &3.84 &4.41 &3.65 \\
\hline
twinGP &4.03&4.65 &3.39 &3.10&3.79 &2.98 &2.88 &3.04 &2.63 &2.88\\
\hline
SV($\boldsymbol x$) &4.62 &5.09 &3.82 &4.03 &4.39&2.77&2.91 &2.89 &2.93 &2.88 \\
\hline
SV($\boldsymbol x$,$t$) &9.27&13.66 &12.48 &12.02&11.86  &3.64 &4.10 &4.37 &4.12 &4.06 \\
\hline
thinned laGP &4.81&5.24 &6.09 &6.59 &5.68 &2.81 &2.97 &3.77 &3.63 &3.30 \\
\hline
thinned twinGP &3.64&3.88 &3.19 &\textbf{2.90} & 3.40&2.62 &\textbf{2.65} &\textbf{2.55} &\textbf{2.45} &\textbf{2.57} \\
\hline
thinned SV &\textbf{3.45}&\textbf{3.59} &\textbf{3.14} &2.94 &\textbf{3.28} &\textbf{2.61} &2.68 &2.79 &2.66 &2.69 \\
\hline
tempGP & 3.39 &  4.35 & 3.45 &3.06 & 3.56& -& -& -&-&-\\
\hline
\end{tabular}
\label{tab:dataset6}
\end{table}

\begin{table}[hb]
\centering
\footnotesize
\caption{\footnotesize Runtime on DSWE Dataset 5 and DSWE  Dataset 6 (in minutes)}
\begin{tabular}{|l|c|c|c|c|c|c|c|c|}
\hline
 & laGP & twinGP & SV($\boldsymbol x$) & SV($\boldsymbol x, t$) & thinned laGP & thinned twinGP & thinned SV &tempGP\\
\hline
Dataset 5 & 31.7 & 0.1 & 1.1 & 1.1 & 24.7 & 2.0 & 3.7 & 83.1\\
\hline
Dataset 6 & 64.4 & 0.3 & 3.4 & 3.4 & 87.9 & 5.9 & 9.5 & 178.7\\
\hline
\end{tabular} \label{tab:runtime_wind}
\end{table}

Through the numerical experiments, we observe that thinned SV delivers the best performance overall, although for most of the cases only marginally better than the second best method, which is the thinned twinGP.  The thinned twinGP is the fastest method among all the thinned version of the GP approximations and only slower than its original version, twinGP.  Both thinned SV and thinned twinGP are considered fast enough; for large datasets, they reduce the runtime by orders of magnitude as compared with the original tempGP method.  When running cross validation on a dataset, including $t$ in SV does help improve its performance on autocorrelated data, but the improvement is not as nearly much as thinned SV improves SV.

\subsection{ Choice of the thinning number \texorpdfstring{$T$}{T}}

Recall that the thinning parameter $T$ is chosen based on Equation~(\ref{thinning_x&y}), which is based on the level of autocorrelation in data. When the data exhibit strong temporal dependence, the partial autocorrelation function (PACF) remains significant across more lags, leading Equation~(\ref{thinning_x&y}) to selects a larger thinning number~$T$. Conversely, when the series is weakly autocorrelated, the PACF decays more rapidly and a smaller~$T$ is obtained. The PACF is computed globally over all valid lagged pairs~$(x_t, x_{t-h})$, which may give greater influence to regions with denser sampling due to their larger number of available pairs in the estimation. In datasets with varying sampling density, such as our wind farm data (Dataset~6), in which a regular month would contain approximately~4{,}320 ten-minute records but we have months with fewer than~2{,}000 due to missing entries, the denser segments exert a stronger effect on the PACF estimate. The resulting global~$T$ therefore tends to thin the densely sampled periods more aggressively. Although this global treatment might overspread the low-frequency regions, such extreme irregularities are rarely encountered in physical measurement systems and do not pose a major concern in our applications.

Here we use thinned SV to see how sensitive the choice made using Equation~(\ref{thinning_x&y}) may be. On the surface, increasing the thinning number increases the number of blocks and reduces the number of data points per block; doing that could accelerate GP computations. 
In reality, computation in thinned SV depends more directly on $m$, which is usually much smaller than the number of data points in a block.  The block size in thinned SV has actually less an impact on computational efficiency. 

One condition limiting the value of $T$ is to ensure that each block has enough data points for choosing the conditioning sets of size $m$. For this reason,
\begin{equation}
\left\lfloor \frac{n}{T_{\text{max}}} \right\rfloor \geq m + 1, \label{eq:condition}
\end{equation}
where the right side is $m+1$ instead of $m$ because one of the data points in the block is the location where we want to find the conditioning set. For a large dataset, like the wind power data, this formula leads to a $T$ still in thousands (recall for the wind power data, $n=45,000$ and $m=30$). 

We believe that this upper limit for $T$ is too loose. The actual $T$ has to be much smaller.  Recall the argument made earlier that each block needs to have a sufficient amount of data so that one can find good enough spatial neighborhoods for the test location $\boldsymbol x^*$.  How many is sufficient? There is no definite answer.  Our empirical experience indicates that it is better for each block has over 1,000 data points.  This puts $T$ to be fewer than 50 if $n=50,000$.  

\begin{figure}[ht]
\centering
\includegraphics[width=.7\textwidth]{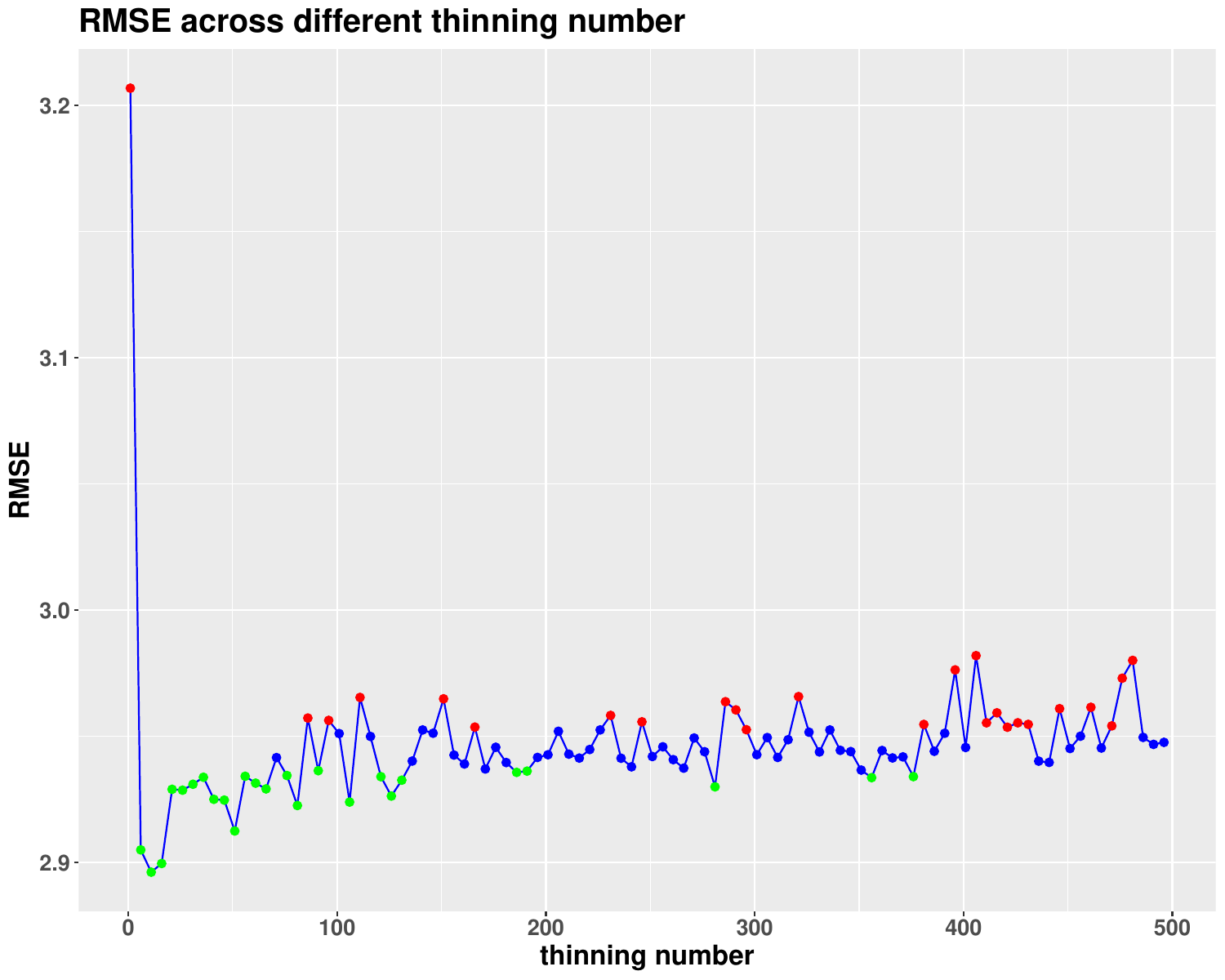} 
\caption{The effect of increasing the thinning number on the out-of-sample RMSE. The 25 highest values are colored in red, and the 25 lowest values are in green.}
\label{fig:rmse_thinningeffect}
\end{figure}

We test thinned SV for a $T$ from 1 to 500 with an increment of 5. The numerical result is the out-of-sample RMSE computed using WT4 data from Dataset 6.  For this example, Equation (\ref{thinning_x&y})'s choice is $T=22$. Figure \ref{fig:rmse_thinningeffect} presents the plot of RMSE. One can observe that the RMSE is lowered by nearly 10\% when the data are thinned, even using a small thinning number.  When $T$ increases, the out-of-sample RMSE shows an unmistakable trend of increasing.  For better illustration, let us color,  among the 100 values, the 25 lowest RMSE in green and the 25 highest RMSE in red.  One red dot is for \( T=1 \), which means no thinning, and the whole training set is used as one block. Other than $T=1$, one sees more green dots towards the end of small $T$'s and more red dots towards the end of large $T$.   When \( T \) is greater than 150, one can still see a few green dots, but the red dots are much more frequent.  In this example, although $T=22$ does not produce the smallest out-of-sample RMSE, its corresponding RMSE is only about 1\% higher than the smallest. 

\subsection{A note on stability}
\begin{figure}[htbp]
\centering
\includegraphics[width=.6\textwidth]{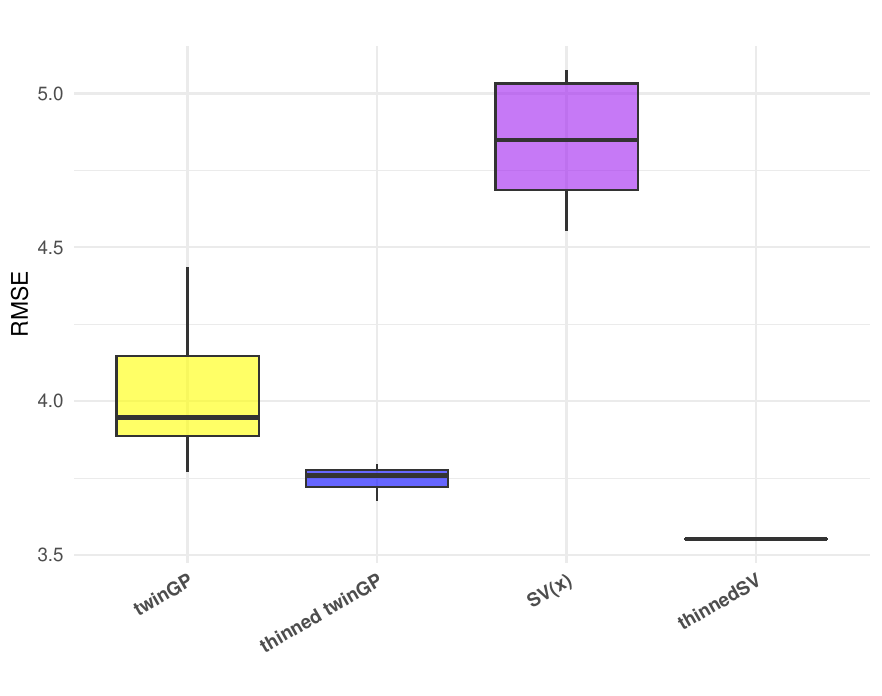} 
\caption{Stability analysis of the leading methods thinned twinGP and thinned SV, along with their unthinned counterparts. The RMSE values are computed over 10 runs, with seeds set from 1 to 10, to assess the consistency of each method.}
\label{fig:stability}
\end{figure}

The numerical experiments in Section \ref{Numerical Experiments} were carried out with a fixed seed for the generation of random numbers. However, we observe that certain methods exhibit sensitivity to the choice of the random seed. We analyze the two leading methods, thinned SV and thinned twinGP, with the random seed values used by them changing from 1 to 10.  We test their unthinned counterparts with the same seed changes. Figure \ref{fig:stability} presents the results using Turbine 1 data in DSWE Dataset 6, although this pattern holds for other datasets. 

We can see that data thinning enhances the numerical stability across methods, with thinned SV demonstrating the highest consistency. The unthinned methods exhibit a greater variability. TwinGP's RMSE changes from 3.77 to 4.44, reflecting a 15\% swing, and SV($\boldsymbol x$)'s RMSE changes from 4.55 to 5.08, a 10\% swing.  By contrast, thinned twinGP is more stable, with its RMSE values between 3.68 and 3.80, a 3\% difference for the range, and thinned SV remains consistent up to at least two decimal places. These findings suggest that data thinning not only improves a GP model's performance on autocorrelated data, it generally enhances numerical stability also.  We want to note that in the numerical experiments conducted earlier, we report the favorable results for the respective unthinned GP approximations.  

\section{Concluding remarks}

This paper addresses the question of how to speed up GP computation on autocorrelated data. We observe that thinning the data before applying GP approximations is critical to achieving good model performances on autocorrelated data. The thinned versions of GP approximations, especially thinned SV and thinned twinGP, improve the resulting GP model's performance remarkably. In contrast, thinned laGP, due to laGP’s longer computational runtime, required the use of a less effective thinning strategy, resulting in inconsistent improvements over the standard laGP. The thinned SV and thinned twinGP methods are readily applicable to data coming from physical systems such as wind farm operational data, climate monitoring, satellite remote sensing, among others.

The fundamental reason behind the speeding up in most of the GP approximations is to use a small subset of data instead of the whole dataset for computation.  This small subset is always selected by using some forms of nearest neighbor strategy, and the nearest neighbor selections in the existing GP approximations are all concerned with spatial neighbors, i.e., based on distances defined on $\boldsymbol x$. When data are autocorrelated, selecting data from a spatial nearest neighborhood almost surely leads to picking data points that are also temporally adjacent, and using which to fit a GP regression model cannot avoid temporal overfitting. 

Thinning the data into disjoint blocks removes autocorrelation among data points in each block. The concern for thinning is that the quality of the spatial neighborhood within each block may suffer. Our research reveals that as long as each block maintains a sufficient size of data points, the benefit of using the thinned data will greatly outweigh a slight deterioration in the quality of spatial neighborhood.  When comparing the prediction made by the best thinned approximation with that from tempGP, the thinned method often delivered a slight improvement in performance. 

The difference in specific nearest neighbor selections associated with the existing GP approximations may explain the different effect of thinning on the respective GP approximation. Recall that thinned SV improves over its original counterpart more than thinned twinGP over its. The approximation methods can be classified into global and local approximations. SV is a global approximation because it uses a product likelihood function to estimate a single set of hyperparameters for one GP model, whereas twinGP is by and large a local approximation which fits localized GP models around test locations.

For the global approximation like SV, we apply thinning to data in the estimation process, but not in the prediction. The use of thinned data in the estimation process is needed for avoiding temporal overfitting but not needed in the prediction process, since the model's hyperparameters would have been estimated already. Using the raw data for selecting the best spatial neighbors for prediction avoids potential quality deterioration in spatial neighborhoods resulting from the use of thinned data.  

The local approximations, on the other hand, use the same set for estimation and prediction.  What this means is that thinning will affect both phases and thus may have a detrimental effect on prediction.  One may ask whether we could also apply thinning \emph{only} in estimation but not in prediction for the local approximations.  It is desirable but not easy to get done, as doing so will tear apart the existing methods, amounting to designing a new method altogether.  By contrast, it is much easier to tailor the Scaled Vecchia, as SV uses two separate conditioning sets.

In summary, the key to having a GP approximation working well for autocorrelated data is to select good spatial neighbors that are temporally far away. Applying GP approximations to the thinned data does this trick successfully. Using the blocks of thinned data entails other benefits.  Prior studies by \citet{Stein2004} and \citet{Vecchia1988} highlight that using nearest neighborhood subsets could lead to suboptimal estimates if the selected points are clustered within narrow regions, thereby limiting the model’s ability to capture broader data variability. \citet{Stein2004} emphasize that subsets should be spatially distributed to avoid bias, as localized subsets risk underestimating uncertainty, ultimately degrading predictive performance. \citet{Stein2004} and \citet{Emery2009} suggest that incorporating more distant neighbors could provide a better approximation of the covariance function. This advice is naturally incorporated in the thinned versions of GP approximation, because when the raw data are thinned into disjoint blocks, the subsequent nearest neighbor selection is forced to choose more distant neighbors.

As a final note, we acknowledge that certain technical treatment inherited from tempGP (e.g., the PACF-based criterion for choosing the thinning parameter $T$) may not be ideal in the cases with highly complex or nonstationary autocorrelation. In the datasets we tested so far, this approach, however simple it may seem, appears to be effective.  Additional research is warranted to look deeper into how to make the choice of $T$ robust and adaptive under more complicated scenarios. 

\vspace{.5cm}
\section*{Supplementary Material}
\vspace{.5cm}
\subsection*{Datasets}
\label{Datasets}
\begin{itemize}
    \item \textbf{Dataset 5 and Dataset 6}: These datasets are publicly available at the \emph{Data Science for Wind Energy} book website. For more information, please visit \href{https://sites.google.com/view/yuding/book-dswe/dswe-datasets}{DSWE Datasets}.

    \item \textbf{Satellite Datasets}: These datasets are publicly available at the LaGP's website. For more information, please visit \href{https://bitbucket.org/gramacylab/tpm/src/master/data/HST/}{LaGP Datasets}.
\end{itemize}


%
%
%

\section*{Funding}

YD’s research was partially supported by National Science Foundation (NSF) Grant CNS--2328395. MK’s research was partially supported by National Science Foundation (NSF) Grant DMS--1953005 and by the Office of the Vice Chancellor for Research and Graduate Education at the University of Wisconsin--Madison with funding from the Wisconsin Alumni Research Foundation. AC’s research was sponsored by the Ocean Energy Safety Institute Consortium (OESIC) through a grant from the U.S.\ Department of the Interior, Bureau of Safety and Environmental Enforcement (BSEE), and the U.S.\ Department of Energy (DOE) and was accomplished under Agreement Number E21AC00000. The views and conclusions contained in this document are those of the authors and should not be interpreted as representing the opinions or policies of the U.S.\ Government. Mention of trade names or commercial products does not constitute their endorsement by the U.S.\ Government.


\bibliographystyle{informs2014} 
\bibliography{bibio} 






  



\end{document}